\documentclass{article}

\usepackage[final]{nips_2017}

\usepackage[skip=0pt]{caption}
\usepackage[skip=0pt]{subcaption}
\usepackage{wrapfig}
\usepackage{multirow}
\usepackage{microtype}
\usepackage{multicol}
\usepackage{tabularx}
\usepackage{graphics}
\usepackage{enumitem}
\usepackage{adjustbox}
\usepackage{newtxtext,newtxmath}
\usepackage{nicefrac}
\usepackage{booktabs}

\usepackage[subtle]{savetrees}
\usepackage{setspace}
\usepackage{pgfplots}
\usetikzlibrary{pgfplots.groupplots}
\pgfplotsset{compat=1.11}
\usepgfplotslibrary{ternary}
\usepackage{tikz}
\usepackage{tkz-graph}
\usetikzlibrary{positioning,automata}
\usetikzlibrary{calc,spy,shapes}
\usetikzlibrary{arrows,decorations.pathmorphing,fit,positioning,patterns}
\pgfplotsset{compat=newest}

\newcommand{\shrink}{\vspace{-1.5ex}}
\newcommand{\sshrink}{\vspace{-.80ex}}

\def\:{\hskip0pt} % \: for hyphenation xxx\:---\:xxx 
\newcommand{\mypar}[1]{\vspace*{-0.8ex}\noindent\textbf{#1}~}
\newcounter{todocnt}

\DeclareFontFamily{U}{mathb}{\hyphenchar\font45}
\DeclareFontShape{U}{mathb}{m}{n}{
<-6> mathb5 <6-7> mathb6 <7-8> mathb7
<8-9> mathb8 <9-10> mathb9
<10-12> mathb10 <12-> mathb12
}{}
\DeclareSymbolFont{mathb}{U}{mathb}{m}{n}

\DeclareMathSymbol{\smalltriangleup} {2}{mathb}{"98}% name to be checked
\DeclareMathSymbol{\smalltriangledown} {2}{mathb}{"99}% name to be checked
\DeclareMathSymbol{\smalltriangleleft} {2}{mathb}{"9A}% name to be checked
\DeclareMathSymbol{\smalltriangleright}{2}{mathb}{"9B}% name to be checked
\DeclareMathSymbol{\blacktriangleup} {2}{mathb}{"9C}% name to be checked
\DeclareMathSymbol{\blacktriangledown} {2}{mathb}{"9D}% name to be checked
\DeclareMathSymbol{\blacktriangleleft} {2}{mathb}{"9E}% name to be checked
\DeclareMathSymbol{\blacktriangleright}{2}{mathb}{"9F}% name to be checked

\usepackage{xspace} % insert space when needed.
\newcommand{\tnet}{\textit{target network}\xspace}

\newcommand{\cnet}{\textit{confidence network}\xspace}

\usepackage[utf8]{inputenc} % allow utf-8 input
\usepackage[T1]{fontenc}    % use 8-bit T1 fonts
\usepackage{hyperref}       % hyperlinks
\usepackage{url}            % simple URL typesetting
\usepackage{booktabs}       % professional-quality tables
\usepackage{amsfonts}       % blackboard math symbols
\usepackage{nicefrac}       % compact symbols for 1/2, etc.
\usepackage{microtype}      % microtypography

\title{Learning to Learn from Weak Supervision \\ by Full Supervision}

\title{Learning to Learn from Weak Supervision \\ 
~~~~~~~~~~~~~~~~~~~~~~~~~~~~~~~~~~~~~~~~~~~
from Full Supervision}

\title{Learning to Learn from Weak Supervision \\ 
~~~~~~~~~~~~~~~~~~~~~~~~~~~~~~~~~~~~~~~~~~~~~~~~
by Full Supervision}  %similar to the title alignment of the "Learning to learn by gradient descent by gradient descent" paper: https://arxiv.org/pdf/1606.04474.pdf :)

\author{
  Mostafa Dehghani \\
  University of Amsterdam\\
  \texttt{dehghani@uva.nl} \\
  \And
  Aliaksei Severyn, Sascha Rothe \\
  Google Research\\
  \texttt{\{severyn,rothe\}@google.com} \\
  \And  
  Jaap Kamps \\
  University of Amsterdam\\
  \texttt{kamps@uva.nl} \\
}

\begin{document}

\maketitle

\begin{abstract}
\shrink
%Making use of weak or noisy signals for training deep neural networks is increasing, in particular for the tasks where an adequate amount of data with true labels is not available. 
% 
%In semi-supervised setting, we can use a large set of data with weak labels to pretrain a neural network and fine tune the parameters with a small amount of data with true labels. However, these two independent stages do not leverage the full capacity of clean information from true labels during pretraining.
%
In this paper, we propose a method for training neural networks when we have a large set of data with weak labels and a small amount of data with true labels. In our proposed model, we train two neural networks: a \tnet, the learner and a \cnet, the meta-learner. 
The \tnet is optimized to perform a given task and is trained using a large set of unlabeled data that are weakly annotated. We propose to control the magnitude of the gradient updates to the \tnet using the scores provided by the second \cnet, which is trained on a small amount of supervised data. Thus we avoid that the weight updates computed from noisy labels harm the quality of the \tnet model.
\shrink
\end{abstract}

\section{Introduction}
\shrink
\label{sec:introduction}
Using weak or noisy supervision is a straightforward approach to increase the size of the training data~\citep{Dehghani:2017:SIGIR, Patrini:2016, Beigman:2009, Zeng:2015, Bunescu:2007}.  The output of heuristic methods can be used as weak or noisy signals along with a small amount of labeled data to train neural networks.
This is usually done by pre-training the network on weak data and fine tuning it with true labels~\citep{Dehghani:2017:SIGIR, Severyn:2015:SIGIR}. 
However, these two independent stages do not leverage the full capacity of information from true labels and using noisy labels of lower quality often brings little to no improvement.
This issue is tackled by noise-aware models where denoising the weak signal is part of the learning process~\citep{Patrini:2016, Sukhbaatar:2014, dehghani2017avoiding}.

In this paper, we propose a method that leverages a small amount of data with true labels along with a large amount of data with weak labels. 
In our proposed method, we train two networks in a multi-task fashion: a \tnet which uses a large set of weakly annotated instances to learn the main task while a \cnet is trained on a small human-labeled set to estimate confidence scores. 
These scores define the magnitude of the weight updates to the \tnet during the back-propagation phase. 
From a meta-learning perspective~\citep{Andrychowicz:2016,Finn2017:ICML,Ravi:2016}, the goal of the \cnet, as the meta-learner, trained jointly with the \tnet, as the learner, is to calibrate the learning rate of the \tnet for each instance in the batch. I.e., the weights $\pmb{w}$ of the \tnet $f_w$ at step $t+1$ are updated as follows:
\begin{equation}
\pmb{w}_{t+1} = \pmb{w}_t - \frac{\eta_t}{b}\sum_{i=1}^b c_{\theta}(x_i, \tilde{y}_i)  \nabla \mathcal{L}(f_{\pmb{w_t}}(x_i), \tilde{y_i})
%+ \nabla \mathcal{R}(\pmb{w_t})
\end{equation}
where $\eta_t$ is the global learning rate, $\mathcal{L}(\cdot)$ is the loss of predicting $\hat{y}=f_w(x_i)$ for an input $x_i$ when the label is $\tilde{y}$; $c_\theta(\cdot)$ is a scoring function learned by the \cnet taking input instance $x_i$ and its noisy label $\tilde{y}_i$. Thus, we can effectively control the contribution to the parameter updates for the \tnet from weakly labeled instances based on how reliable their labels are according to the \cnet (learned on a small supervised data).

Our approach is similar to~\citep{Andrychowicz:2016}, where a separate recurrent neural network called \textit{optimizer} learns to predict an optimal update rule for updating parameters of the \tnet. 
The optimizer receives a gradient from the \tnet and outputs the adjusted gradient matrix. As the number of parameters in modern neural networks is typically on the order of millions the gradient matrix becomes too large to feed into the optimizer, so the approach presented in~\citep{Andrychowicz:2016} is applied to very small models. 
In contrast, our approach leverages additional weakly labeled data where we use the \cnet to predict per-instance scores that calibrate gradient updates for the \tnet.

\begin{figure*}[!t]%
    \centering
    \begin{subfigure}[t]{0.49\textwidth}
        \centering
        \includegraphics[width=0.95\textwidth]{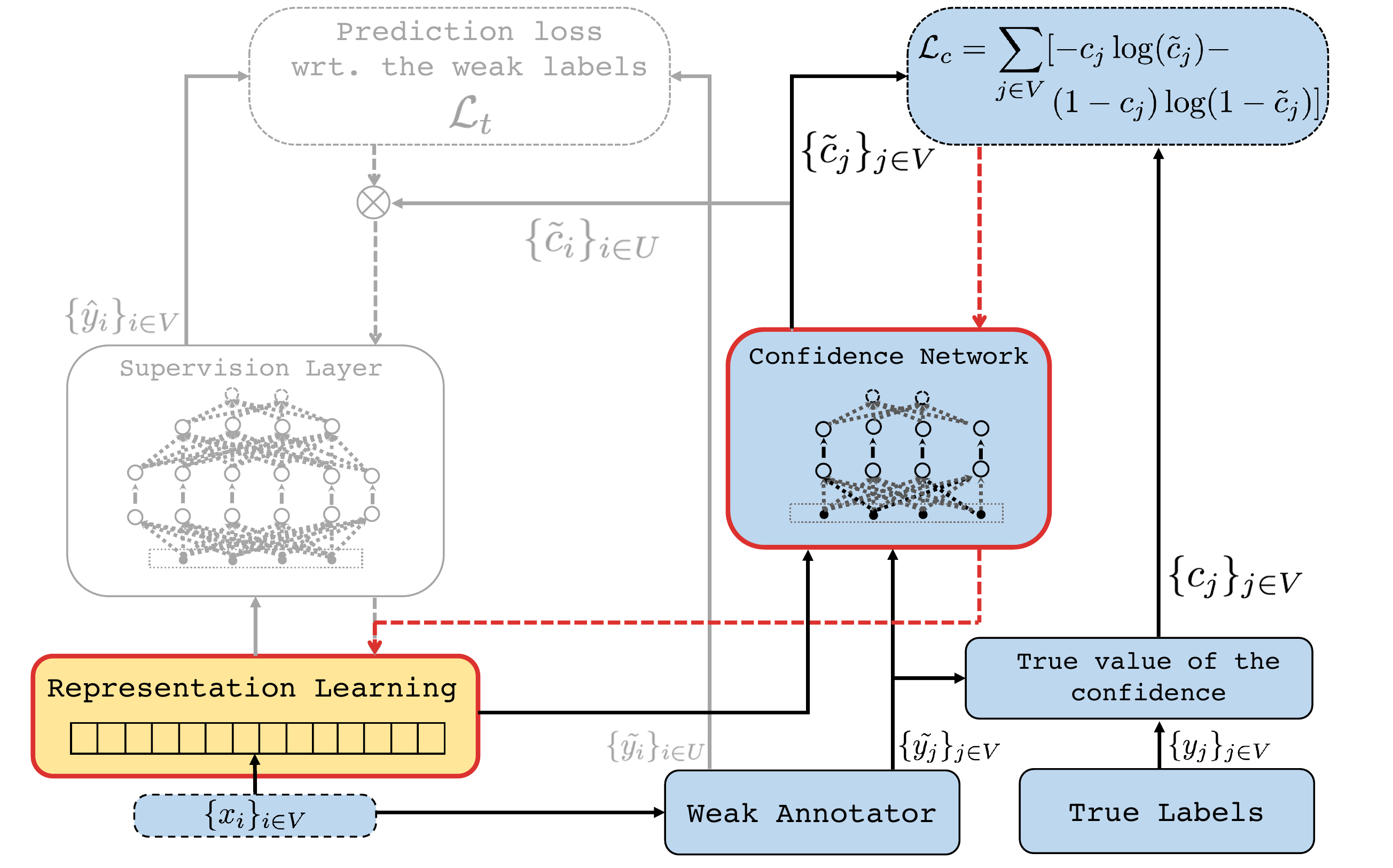}
        % \vspace{-2pt}
        \caption{\label{fig:train_u}\tiny{Full Supervision Mode: Training on batches of data with true labels.}}
        % \vspace{-1pt}
    \end{subfigure}%
    ~
    \begin{subfigure}[t]{0.49\textwidth}
        \centering
        \includegraphics[width=0.95\textwidth]{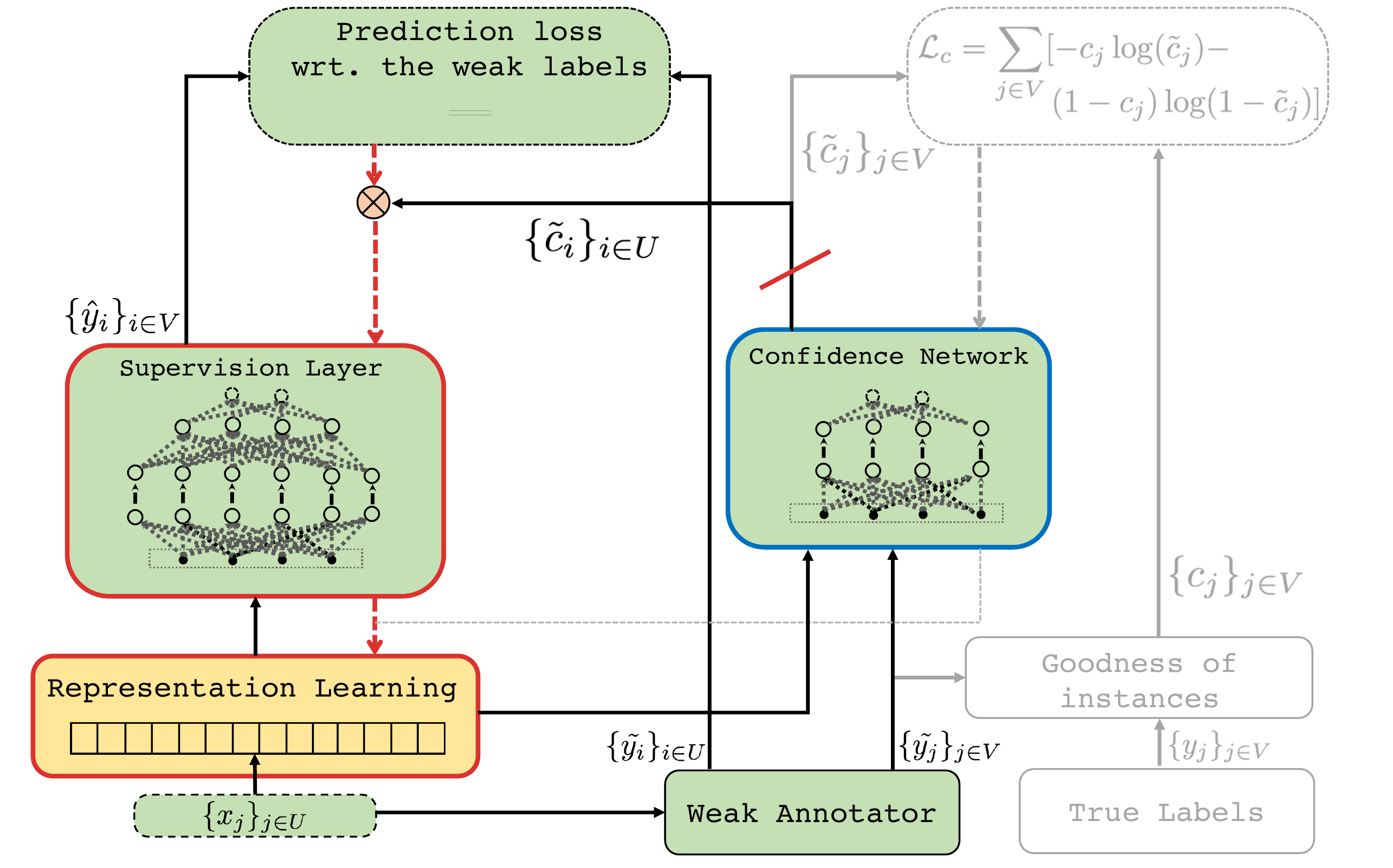}
        % \vspace{-2pt}
        \caption{\label{fig:train_v}\tiny{Weak Supervision Mode: Training on batches of data with weak labels.}}
        % \vspace{-1pt}
    \end{subfigure}%
    \caption{\fontsize{8}{9}\selectfont{Our proposed multi-task network for learning a target task using a large amount of weakly labeled data and a small amount of data with true labels. 
    Faded parts of the network are disabled during the training in the corresponding mode. Red-dotted arrows show gradient propagation. Parameters of the parts of the network in red frames get updated in the backward pass, while parameters of the network in blue frames are fixed during the training.}}
    \label{fig:model}
    \vspace{-20pt}
\end{figure*}
Our setup requires running a weak annotator to label a large amount of unlabeled data, which is done at pre-processing time. For many tasks, it is possible to use a simple heuristic to generate weak labels. This set is then used to train the \tnet.  
In contrast, a small human-labeled set is used to train the \cnet, which estimates how good the weak annotations are, i.e. controls the effect of weak labels on updating the parameters of the \tnet.
This helps to alleviate updates from instances with unreliable labels that may corrupt the \tnet.

In this paper, we study our approach on sentiment classification task.Our experimental results suggest that the proposed method is more effective in leveraging large amounts of weakly labeled data compared to traditional fine-tuning.
We also show that explicitly controlling the \tnet weight updates with the \cnet leads to faster convergence.
\shrink
\section{The Proposed Method}
\shrink
In the following, we describe our recipe for training neural networks, in a scenario where along with a small human-labeled training set a large set of weakly labeled instances is leveraged.
Formally, given a set of unlabeled training instances, we run a weak annotator to generate noisy labels.
This gives us the training set $U$. It consists of \emph{tuples} of training instances $x_i$ and their weak labels $\tilde{y}_i$, i.e. $U=\{(x_i, \tilde{y}_i),\ldots\}$. 
For a small set of training instances with true labels, we also apply the weak annotator to generate weak labels.
This creates the training set $V$, consisting of \emph{triplets} of training instances $x_j$, their weak labels $\tilde{y}_j$, and their true labels $y_j$, i.e. $V=\{(x_j,\tilde{y}_j,y_j),\ldots\}$. 
We can generate a large amount of training data $U$ at almost no cost using the weak annotator.
In contrast, we have only a limited amount of data with true labels, i.e. $|V|<<|U|$. 

% \subsection{General Architecture}
% \label{sec:generalarchitecture}
In our proposed framework we train a multi-task neural network that jointly learns the confidence score of weak training instances and the main task using controlled supervised signals.
The high-level representation of the model is shown in Figure~\ref{fig:model}: it comprises two neural networks, namely the \cnet and the \tnet. 
The goal of the \cnet is to \emph{estimate the confidence score} $\tilde{c}_j$ of training instances. It is learned on triplets from training set $V$: input $x_j$, its weak label $\tilde{y}_j$, and its true label $y_j$.
The score $\tilde{c}_j$ is then used to control the effect of weakly annotated training instances on updating the parameters of the \tnet. 

%in its backward pass during back propagation.

The \tnet is in charge of \emph{handling the main task} we want to learn. Given the data instance, $x_i$ and its weak label $\tilde{y}_i$ from the training set $U$, the \tnet aims to predict the label $\hat{y}_i$. 
The \tnet parameter updates are based on noisy labels assigned by the weak annotator, but the magnitude of the gradient update is based on the output of the \cnet. 

Both networks are trained in a multi-task fashion alternating between the \emph{full supervision} and the \emph{weak supervision} mode.  
In the \emph{full supervision} mode, the parameters of the \cnet get updated using batches of instances from training set $V$.  
As depicted in Figure~\ref{fig:train_v}, each training instance is passed through the representation layer mapping inputs to vectors. These vectors are concatenated with their corresponding weak labels $\tilde{y}_j$. %generated by the weak annotator.
The \cnet then estimates $\tilde{c}_j$, which is the probability of taking data instance $j$ into account for training the \tnet.

In the \emph{weak supervision} mode the parameters of the \tnet are updated using training set $U$.
As shown in Figure~\ref{fig:train_u}, each training instance is passed through the same representation learning layer and is then processed by the supervision layer which is a part of the \tnet predicting the label for the main task. 
We also pass the learned representation of each training instance along with its corresponding label generated by the weak annotator to the \cnet to estimate the \emph{confidence score} of the training instance, i.e. $\tilde{c}_i$. 
%The confidence score of the instance is then used as the weight of the loss to control the gradient and consequently the amount of parameter updates in the backward pass of the based network. 
The confidence score is computed for each instance from set $U$. These confidence scores are used to weight the gradient updating the \tnet parameters during back-propagation.
It is noteworthy that the representation layer is shared between both networks, so the \cnet can benefit from the largeness of set $U$ and the \tnet can utilize the quality of set $V$. 
\shrink
\subsection{Model Training}
\shrink
\label{sec:modeltraining}
Our optimization objective is composed of two terms: (1) the \cnet loss $\mathcal{L}_c$, which captures the quality of the output from the \cnet and (2) the \tnet loss $\mathcal{L}_t$, which expresses the quality for the main task. 
%The parameters of both networks are updated in alternating fashion with a fixed ratio. %will be repeated

Both networks are trained by alternating between the \emph{weak supervision} and the \emph{full supervision} mode.
In the \emph{full supervision} mode, the parameters of the \cnet are updated using training instance drawn from training set $V$. We use cross-entropy loss function for the \cnet to capture the difference between the predicted confidence score of instance $j$, i.e. $\tilde{c}_j$ and the target score $c_j$:
% \begin{equation}
% \nonumber
$\mathcal{L}_c = \sum_{j\in V} -  c_j \log(\tilde{c}_j) - (1-c_j) \log(1-\tilde{c}_j)$,
% \end{equation}
The target score $c_j$ is calculated based on the difference of the true and weak labels with respect to the main task.
In the \emph{weak supervision} mode, the parameters of the \tnet are updated using training instances from $U$. We use a weighted loss function, $\mathcal{L}_t$, to capture the difference between the predicted label $\hat{y}_i$ by the \tnet and target label $\tilde{y}_i$:
% \begin{equation}
% \nonumber
$\mathcal{L}_t = \sum_{i\in U} \tilde{c}_i \mathcal{L}_i$,
% \end{equation}
where $\mathcal{L}_i$ is the task-specific loss on training instance $i$ and $\tilde{c}_i$ is the confidence score of the weakly annotated instance $i$, estimated by the \cnet.
Note that $\tilde{c}_i$ is treated as a constant during the weak supervision mode and there is no gradient propagation to the \cnet in the backward pass (as depicted in Figure~\ref{fig:train_u}). 

We minimize two loss functions jointly by randomly alternating between full and weak supervision modes (for example, using a 1:10 ratio).
During training and based on the chosen supervision mode, we sample a batch of training instances from $V$ with replacement or from $U$ without replacement (since we can generate as much train data for set $U$). 
%

%The key point here is that the ``main task'' and ``confidence scoring'' task are always defined to be close tasks and sharing representation will benefit the confidence network as an implicit data augmentation to compensate the small amount of data with true labels. Besides, we noticed that updating the representation layer with respect to the loss of the other network acts as a regularization which helps generalization for both target and confidence network since we try to capture all tasks (which are related tasks) and less chance for overfitting.

% updating the parameters of the confidence network during full supervision mode but it does not work.  The main reason is that the gradient from the loss in the full-supervision mode is with respect to the error of the main task, while the task we want to learn in the confidence network is to estimating how good is a weakly labeled training instances.  Nevertheless, the confidence network benefits from learned information from these updates by using the shared representation layer. 

% We also investigated other possible setups like updating the parameters of the base network using also data with true labels.
% Instead of using alternating sampling, we can train the \tnet using controlled weak supervision signals after the \cnet is fully trained.
% As shown in the experiments the architecture and training strategy described above provide the best performance.
%
\shrink
\section{Experiments}
\shrink
In this section, we apply our method to \emph{sentiment classification} task. 
This task aims to identify the sentiment (e.g., positive, negative, or neutral) underlying an individual sentence.
Our \tnet is a convolutional model similar to~\citep{Deriu:2017, Severyn:2015:SIGIR, Severyn:2015:SemEval,Deriu2016:SemEval}. In this model, the \textit{Representation Learning Layer} learns to map the input sentence $s$ to a dense vector as its representation. The inputs are first passed through an embedding layer mapping the sentence to a matrix $S \in \mathbb{R}^{m \times |s|}$, followed by a series of 1d convolutional layers with max-pooling.
The \textit{Supervision Layer} is a feed-forward neural network with softmax instead as the output layer which returns the probability distribution over all three classes. 
As the the \textit{Weak Annotator}, for the sentiment classification task is a simple unsupervised lexicon-based method~\citep{Hamdan:2013,Kiritchenko:2014}, which averages over predefined sentiment score of words~\citep{Gaccianella:2010} in the sentence. 
More details about the sentiment classification model and the experimental setups are provided in Appendix~\ref{app:sent} and Appendix~\ref{app:setup}, respectively.
In the following, we briefly introduce our baselines, dataset we have used, and present results of the experiments.

\mypar{Baselines.}We evaluate the performance of our method compared to the following baselines:
(\textbf{WA}) Weak Annotator, i.e. the unsupervised method that we used for annotating the unlabeled data.
(\textbf{WSO}) Weak Supervision Only, i.e. the \tnet trained only on weakly labeled data.
(\textbf{FSO}) Full Supervision Only, i.e. the \tnet trained only on true labeled data.
(\textbf{WS+FT}) Weak Supervision + Fine Tuning, i.e. the \tnet trained on the weakly labeled data and fine tuned on true labeled data.
(\textbf{NLI}) New Label Inference~\citep{Veit:2017} is similar to our proposed neural architecture inspired by the teacher-student paradigm~\citep{Hinton:2015}, but instead of having the \cnet to predict the ``confidence score'' of the training instance, there is a \emph{label generator network} which is trained on set $V$ to map the weak labels of the instances in $U$ to the \emph{new labels}. The new labels are then used as the target for training the \tnet.
(\textbf{L2LWS$_\text{ST}$}) Our model with different training setup: Separate Training, i.e. we consider the \cnet as a separate network, without sharing the representation learning layer, and train it on set $V$. We then train the \tnet on the controlled weak supervision signals.
(\textbf{L2LWS}) Learning to Learn from Weak Supervision with Joint Training is our proposed neural architecture in which we jointly train the \tnet and the \cnet by alternating batches drawn from sets $V$ and $U$ (as explained in Section~\ref{sec:modeltraining}). 

\mypar{Data.}
For train/test our model, we use SemEval-13 SemEval-14, SemEval-15, twitter sentiment classification task. We use a large corpus containing 50M tweets collected during two months as unblabled set.

\newcommand{\ps}{$^\blacktriangleup$}
\newcommand{\ns}{$^\smalltriangledown$}
\newcommand{\fs}{$^{~}$}

\begin{figure}[!t]
    \adjustbox{valign=t}{\begin{minipage}[b]{0.40\linewidth}
            \centering
            \captionof{table}{\fontsize{8}{9}\selectfont{Performance of the baseline models as well as our proposed method on different datasets in terms of Macro-F1.  \ps or\ns indicates that the improvements or degradations with respect to weak supervision only ({\small{WSO}}) are statistically significant, at the 0.05 level using the paired two-tailed t-test.}}
            \label{tbl_main_sent}
            \vspace{10pt}
            \begin{adjustbox}{max width=\columnwidth}
            \begin{tabular}{l c c}
            \toprule
            Method & SemEval-14 & SemEval-15
            \\ \midrule \\ [-0.8em]
            \small{WA$_\text{Lexicon}$} & 0.5141 & 0.4471
            \\  \midrule  \\[-0.8em]
            \small{WSO} & 0.6719\fs & 0.5606\fs 
            \\  \\[-0.8em]
            \small{FSO} & 0.6307\fs & 0.5811\fs
            \\ \midrule  \\[-0.8em]
            \small{WS+FT} & 0.7080\ps & 0.6441\ps
            % \\
            % \small{WS+SFT} & 0.6875\fs & 0.6193\ps
            % \\
            % \small{WS+RFT} & 0.6932\fs & 0.6102\ps
            \\ \midrule  \\[-0.8em]
            \small{NLI} & 0.7113\ps & 0.6433\ps
            % \\
            % \small{CWS$_\text{JT}^+$} & 0.7310\ps & 0.6551\ps
            \\ \midrule  \\[-0.8em]
            \small{L2LWS$_\text{ST}$} & 0.7183\ps & 0.6501\ps
            \\   \\[-0.8em]
            \small{L2LWS} & \textbf{0.7362}\ps & \textbf{0.6626}\ps
            \\ \midrule  \\[-0.8em]
            \small{SemEval$^\text{1th}$} & 0.7162\ps & 0.6618\ps
            \\\bottomrule
            \end{tabular}
            \end{adjustbox}
    \end{minipage}
    }
    \hfill
    \adjustbox{valign=t}{\begin{minipage}[b]{0.55\linewidth}
        \centering
            \centering
            \includegraphics[width=0.95\linewidth]{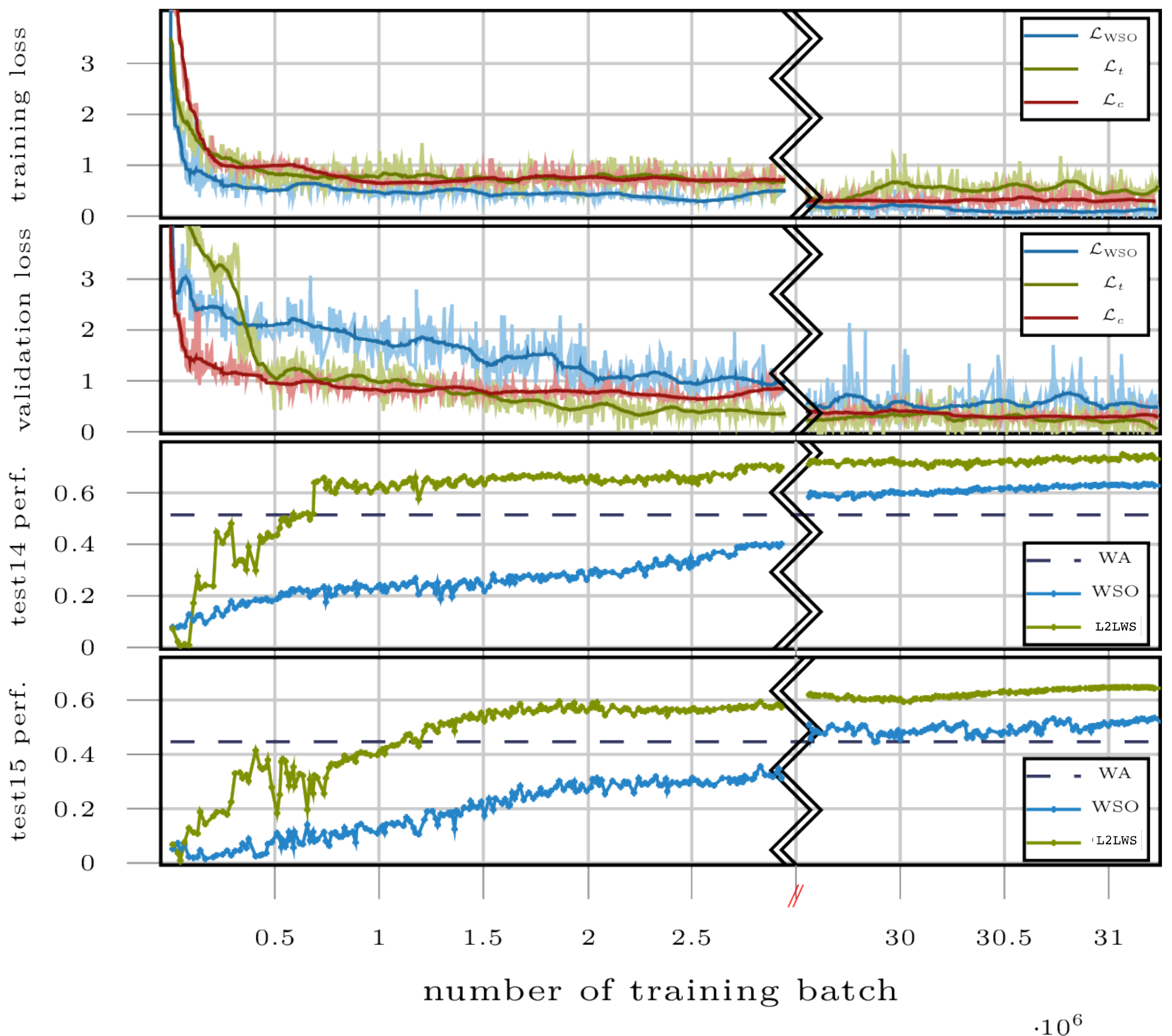}
            % \vspace{-20pt}
            \captionof{figure}{\fontsize{8}{9}\selectfont{Loss of the \tnet ($\mathcal{L}_t$) and the \cnet ($\mathcal{L}_c$) compared to the loss of WSO ($\mathcal{L}_\text{WSO}$) on training/validation set and performance of L2LWS, WSO, and WA on test sets with respect to different amount of training data on sentiment classification.}}
            \label{fig:plot}
    \end{minipage}
    }
    \vspace{-15pt}
\end{figure}
\mypar{Results and Discussion.}
We report the official SemEval metric, Macro-F1, in Table~\ref{tbl_main_sent}.  Based on the results,  L2LWS provides a significant boost on the performance over all datasets.
Typical fine tuning, i.e. WS+FT, leads to improvement over weak supervision only. 
The performance of NLI is worse than L2LWS as learning a mapping from imperfect labels to accurate labels and training the \tnet on new labels is essentially harder than learning to filter out the noisy labels, hence needs a lot of supervised data.
L2LWS$_\text{ST}$ performs worse than L2LWS since the training data $V$ is not enough to train a high-quality \cnet without taking advantage of the shared representation that can be learned from the vast amount of weakly annotated data in $U$. We also noticed that this strategy leads to a slow convergence compared to WSO.
Besides the general baselines, we also report the best performing systems, which are also convolution-based models (\citep{Rouvier:2016} on SemEval-14; \citep{Deriu2016:SemEval} on SemEval-15). Our proposed model outperforms the best systems.

Controlling the effect of supervision to train neural networks not only improves the performance, but also provides the network with more solid signals which speeds up the training process. 
Figure~\ref{fig:plot} illustrates the training/validation loss for both networks, compared to the loss of training the \tnet with weak supervision, along with their performance on test sets, with respect to different amounts of training data for the sentiment classification task.
As shown, training, $\mathcal{L}_t$ is higher than $\mathcal{L}_\text{WSO}$, but the target labels with respect of which the loss is calculated, are weak, so regardless overfitting problem and lack of generalization, a very low loss means fitting the imperfection of the weak data. 
However, $\mathcal{L}_t$ in the validation decreases faster than $\mathcal{L}_\text{WSO}$ and compared to WSO, the performance of L2LWS on both test sets increases quickly and L2LWS passes the performance of the weak annotator by seeing fewer instances annotated by WA.

\sshrink
\shrink
\section{Conclusion}
\shrink
% Training neural networks using large amounts of weakly annotated data is an attractive approach in scenarios where an adequate amount of data with true labels is not available.
In this paper, we propose a neural network architecture that unifies learning to estimate the confidence score of weak annotations and training neural networks with controlled weak supervision.
We apply the model to the sentiment classification task, and empirically verify that the proposed model speeds up the training process and obtains more accurate results. 
%As a promising future direction, we are going to 
%understand to which extent using weak annotations has the potential of training high-quality models with neural networks and 
%understand the exact conditions under which our proposed method works.

% \bibliographystyle{plain}
% \bibliographystyle{unsrt}
\bibliographystyle{plainnat}
\bibliography{ref} 

\begin{thebibliography}{27}
\providecommand{\natexlab}[1]{#1}
\providecommand{\url}[1]{\texttt{#1}}
\expandafter\ifx\csname urlstyle\endcsname\relax
  \providecommand{\doi}[1]{doi: #1}\else
  \providecommand{\doi}{doi: \begingroup \urlstyle{rm}\Url}\fi

\bibitem[Abadi et~al.(2015)]{tensorflow2015-whitepaper}
Mart\'{\i}n Abadi et~al.
\newblock {TensorFlow}: Large-scale machine learning on heterogeneous systems,
  2015.
\newblock URL \url{http://tensorflow.org/}.
\newblock Software available from tensorflow.org.

\bibitem[Andrychowicz et~al.(2016)Andrychowicz, Denil, Gomez, Hoffman, Pfau,
  Schaul, and de~Freitas]{Andrychowicz:2016}
Marcin Andrychowicz, Misha Denil, Sergio Gomez, Matthew~W Hoffman, David Pfau,
  Tom Schaul, and Nando de~Freitas.
\newblock Learning to learn by gradient descent by gradient descent.
\newblock In \emph{Advances in Neural Information Processing Systems}, pages
  3981--3989, 2016.

\bibitem[Baccianella et~al.(2010)Baccianella, Esuli, and
  Sebastiani]{Gaccianella:2010}
Stefano Baccianella, Andrea Esuli, and Fabrizio Sebastiani.
\newblock Sentiwordnet 3.0: An enhanced lexical resource for sentiment analysis
  and opinion mining.
\newblock In \emph{LREC}, volume~10, pages 2200--2204, 2010.

\bibitem[Beigman and Klebanov(2009)]{Beigman:2009}
Eyal Beigman and Beata~Beigman Klebanov.
\newblock Learning with annotation noise.
\newblock In \emph{Proceedings of the Joint Conference of the 47th Annual
  Meeting of the ACL and the 4th International Joint Conference on Natural
  Language Processing of the AFNLP: Volume 1-Volume 1}, pages 280--287.
  Association for Computational Linguistics, 2009.

\bibitem[Bunescu and Mooney(2007)]{Bunescu:2007}
Razvan Bunescu and Raymond Mooney.
\newblock Learning to extract relations from the web using minimal supervision.
\newblock In \emph{ACL}, 2007.

\bibitem[Dehghani et~al.(2017{\natexlab{a}})Dehghani, Severyn, Rothe, and
  Kamps]{dehghani2017avoiding}
Mostafa Dehghani, Aliaksei Severyn, Sascha Rothe, and Jaap Kamps.
\newblock Avoiding your teacher's mistakes: Training neural networks with
  controlled weak supervision.
\newblock \emph{arXiv preprint arXiv:1711.00313}, 2017{\natexlab{a}}.

\bibitem[Dehghani et~al.(2017{\natexlab{b}})Dehghani, Zamani, Severyn, Kamps,
  and Croft]{Dehghani:2017:SIGIR}
Mostafa Dehghani, Hamed Zamani, Aliaksei Severyn, Jaap Kamps, and W.~Bruce
  Croft.
\newblock Neural ranking models with weak supervision.
\newblock In \emph{SIGIR'17}, 2017{\natexlab{b}}.

\bibitem[Deriu et~al.(2016)Deriu, Gonzenbach, Uzdilli, Lucchi, De~Luca, and
  Jaggi]{Deriu2016:SemEval}
Jan Deriu, Maurice Gonzenbach, Fatih Uzdilli, Aurelien Lucchi, Valeria De~Luca,
  and Martin Jaggi.
\newblock Swisscheese at semeval-2016 task 4: Sentiment classification using an
  ensemble of convolutional neural networks with distant supervision.
\newblock \emph{Proceedings of SemEval}, pages 1124--1128, 2016.

\bibitem[Deriu et~al.(2017)Deriu, Lucchi, De~Luca, Severyn, M{\"u}ller,
  Cieliebak, Hofmann, and Jaggi]{Deriu:2017}
Jan Deriu, Aurelien Lucchi, Valeria De~Luca, Aliaksei Severyn, Simon
  M{\"u}ller, Mark Cieliebak, Thomas Hofmann, and Martin Jaggi.
\newblock Leveraging large amounts of weakly supervised data for multi-language
  sentiment classification.
\newblock In \emph{Proceedings of the 26th international International World
  Wide Web Conference (WWW'17)}, pages 1045--1052, 2017.

\bibitem[Desautels et~al.(2014)Desautels, Krause, and Burdick]{Desautels:2014}
Thomas Desautels, Andreas Krause, and Joel~W Burdick.
\newblock Parallelizing exploration-exploitation tradeoffs in gaussian process
  bandit optimization.
\newblock \emph{Journal of Machine Learning Research}, 15\penalty0
  (1):\penalty0 3873--3923, 2014.

\bibitem[Finn et~al.(2017)Finn, Abbeel, and Levine]{Finn2017:ICML}
Chelsea Finn, Pieter Abbeel, and Sergey Levine.
\newblock Model-agnostic meta-learning for fast adaptation of deep networks.
\newblock In \emph{ICML}, 2017.

\bibitem[Hamdan et~al.(2013)Hamdan, B{\'e}chet, and Bellot]{Hamdan:2013}
Hussam Hamdan, Frederic B{\'e}chet, and Patrice Bellot.
\newblock Experiments with dbpedia, wordnet and sentiwordnet as resources for
  sentiment analysis in micro-blogging.
\newblock In \emph{Second Joint Conference on Lexical and Computational
  Semantics (* SEM)}, volume~2, pages 455--459, 2013.

\bibitem[Hinton et~al.(2015)Hinton, Vinyals, and Dean]{Hinton:2015}
Geoffrey Hinton, Oriol Vinyals, and Jeff Dean.
\newblock Distilling the knowledge in a neural network.
\newblock \emph{arXiv preprint arXiv:1503.02531}, 2015.

\bibitem[Kingma and Ba(2014)]{Kingma:2014}
Diederik Kingma and Jimmy Ba.
\newblock Adam: A method for stochastic optimization.
\newblock \emph{arXiv preprint arXiv:1412.6980}, 2014.

\bibitem[Kiritchenko et~al.(2014)Kiritchenko, Zhu, and
  Mohammad]{Kiritchenko:2014}
Svetlana Kiritchenko, Xiaodan Zhu, and Saif~M Mohammad.
\newblock Sentiment analysis of short informal texts.
\newblock \emph{Journal of Artificial Intelligence Research}, 50:\penalty0
  723--762, 2014.

\bibitem[Nakov et~al.(2016)Nakov, Ritter, Rosenthal, Sebastiani, and
  Stoyanov]{Nakov:2016}
Preslav Nakov, Alan Ritter, Sara Rosenthal, Fabrizio Sebastiani, and Veselin
  Stoyanov.
\newblock Semeval-2016 task 4: Sentiment analysis in twitter.
\newblock \emph{Proceedings of SemEval}, pages 1--18, 2016.

\bibitem[Patrini et~al.(2016)Patrini, Rozza, Menon, Nock, and Qu]{Patrini:2016}
Giorgio Patrini, Alessandro Rozza, Aditya Menon, Richard Nock, and Lizhen Qu.
\newblock Making neural networks robust to label noise: a loss correction
  approach.
\newblock \emph{arXiv preprint arXiv:1609.03683}, 2016.

\bibitem[Ravi and Larochelle(2016)]{Ravi:2016}
Sachin Ravi and Hugo Larochelle.
\newblock Optimization as a model for few-shot learning.
\newblock In \emph{ICLR}, 2016.

\bibitem[Rosenthal et~al.(2015)Rosenthal, Nakov, Kiritchenko, Mohammad, Ritter,
  and Stoyanov]{rosenthal:2015}
Sara Rosenthal, Preslav Nakov, Svetlana Kiritchenko, Saif~M Mohammad, Alan
  Ritter, and Veselin Stoyanov.
\newblock Semeval-2015 task 10: Sentiment analysis in twitter.
\newblock In \emph{Proceedings of the 9th international workshop on semantic
  evaluation (SemEval 2015)}, pages 451--463, 2015.

\bibitem[Rouvier and Favre(2016)]{Rouvier:2016}
Mickael Rouvier and Benoit Favre.
\newblock Sensei-lif at semeval-2016 task 4: Polarity embedding fusion for
  robust sentiment analysis.
\newblock \emph{Proceedings of SemEval}, pages 202--208, 2016.

\bibitem[Severyn and Moschitti(2015{\natexlab{a}})]{Severyn:2015:SIGIR}
Aliaksei Severyn and Alessandro Moschitti.
\newblock Twitter sentiment analysis with deep convolutional neural networks.
\newblock In \emph{Proceedings of the 38th International ACM SIGIR Conference
  on Research and Development in Information Retrieval}, pages 959--962. ACM,
  2015{\natexlab{a}}.

\bibitem[Severyn and Moschitti(2015{\natexlab{b}})]{Severyn:2015:SemEval}
Aliaksei Severyn and Alessandro Moschitti.
\newblock Unitn: Training deep convolutional neural network for twitter
  sentiment classification.
\newblock In \emph{Proceedings of the 9th International Workshop on Semantic
  Evaluation (SemEval 2015), Association for Computational Linguistics, Denver,
  Colorado}, pages 464--469, 2015{\natexlab{b}}.

\bibitem[Srivastava et~al.(2014)Srivastava, Hinton, Krizhevsky, Sutskever, and
  Salakhutdinov]{Srivastava:2014}
Nitish Srivastava, Geoffrey Hinton, Alex Krizhevsky, Ilya Sutskever, and Ruslan
  Salakhutdinov.
\newblock Dropout: A simple way to prevent neural networks from overfitting.
\newblock \emph{J. Mach. Learn. Res.}, 15\penalty0 (1):\penalty0 1929--1958,
  2014.

\bibitem[Sukhbaatar et~al.(2014)Sukhbaatar, Bruna, Paluri, Bourdev, and
  Fergus]{Sukhbaatar:2014}
Sainbayar Sukhbaatar, Joan Bruna, Manohar Paluri, Lubomir Bourdev, and Rob
  Fergus.
\newblock Training convolutional networks with noisy labels.
\newblock \emph{arXiv preprint arXiv:1406.2080}, 2014.

\bibitem[Tang(2016)]{tang2016:tflearn}
Yuan Tang.
\newblock Tf.learn: Tensorflow's high-level module for distributed machine
  learning.
\newblock \emph{arXiv preprint arXiv:1612.04251}, 2016.

\bibitem[Veit et~al.(2017)Veit, Alldrin, Chechik, Krasin, Gupta, and
  Belongie]{Veit:2017}
Andreas Veit, Neil Alldrin, Gal Chechik, Ivan Krasin, Abhinav Gupta, and Serge
  Belongie.
\newblock Learning from noisy large-scale datasets with minimal supervision.
\newblock In \emph{The Conference on Computer Vision and Pattern Recognition},
  2017.

\bibitem[Zeng et~al.(2015)Zeng, Liu, Chen, and Zhao]{Zeng:2015}
Daojian Zeng, Kang Liu, Yubo Chen, and Jun Zhao.
\newblock Distant supervision for relation extraction via piecewise
  convolutional neural networks.
\newblock In \emph{EMNLP}, pages 1753--1762, 2015.

\end{thebibliography}

\newpage
\section*{Appendices}
\appendix 
\section{Sentiment Classification Model}
\label{app:sent}
In the sentiment classification task, we aim to identify the sentiment (e.g., positive, negative, or neutral) underlying an individual sentence.
The model we used as the \tnet is a convolutional model similar to~\citep{Deriu:2017, Severyn:2015:SIGIR, Severyn:2015:SemEval,Deriu2016:SemEval}. 

\begin{wrapfigure}{r}{0.31\textwidth}
    \vspace{-10pt}
    \centering
         \includegraphics[width=0.3\textwidth]{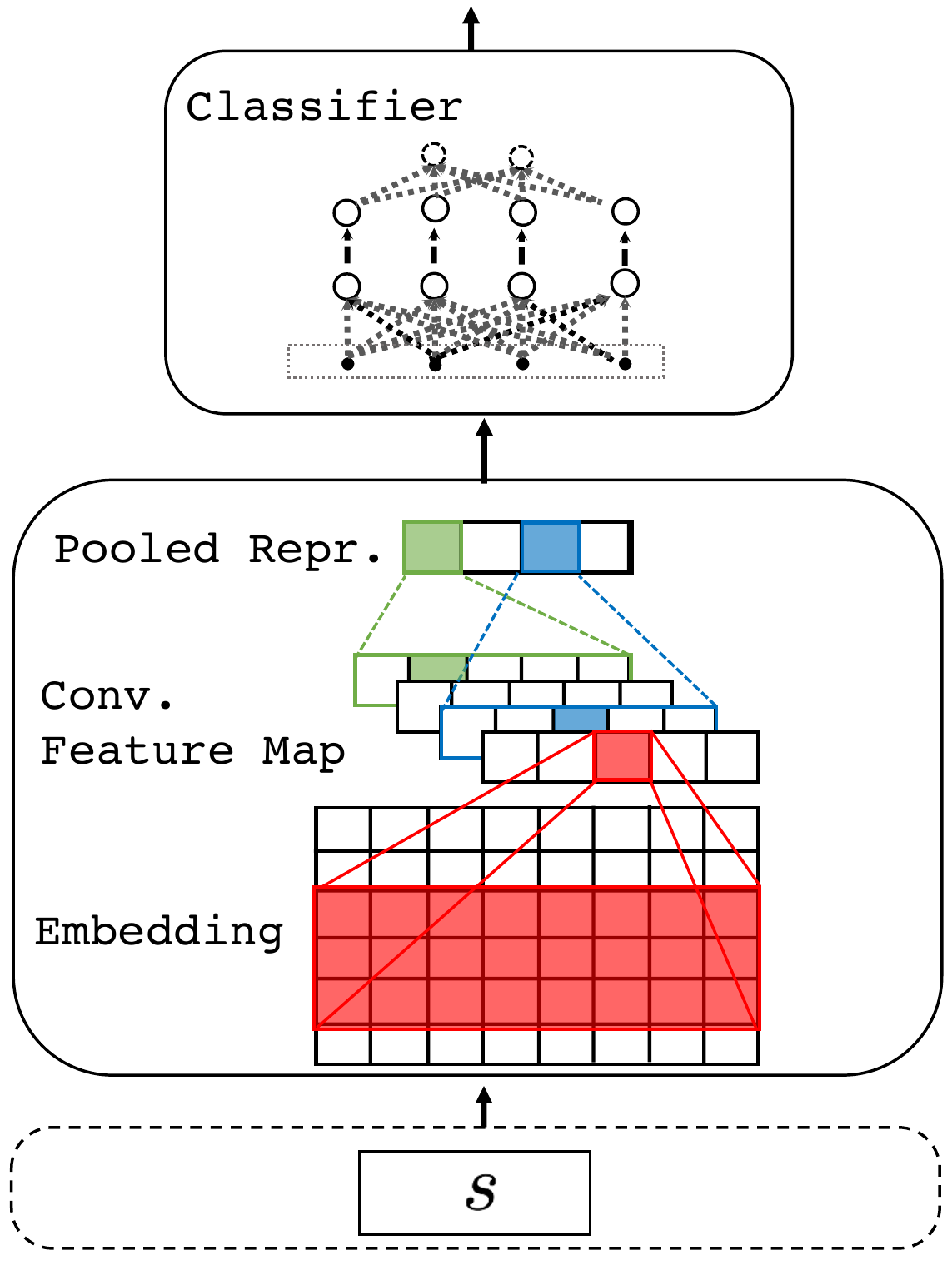}
    \caption{\fontsize{8}{9}\selectfont{The \tnet for the sentiment classification task.}}
    \label{fig:sentiment}   
    \vspace{-15pt}
\end{wrapfigure}

Each training instance $x$ consists of a sentence $s$ and its sentiment label $\tilde{y}$. The architecture of the \tnet is illustrated in Figure~\ref{fig:sentiment}. Here we describe the setup of the target network, i.e. description of the representation learning layer and the supervision layer.

\mypar{The Representation Learning Layer} learns a representation for the input sentence $s$ and is shared between the \tnet and \cnet.
It consists of an embedding function $\varepsilon: \mathcal{V} \rightarrow \mathbb{R}^{m}$, where $\mathcal{V}$ denotes the vocabulary set and $m$ is the number of embedding dimensions.

This function maps the sentence to a matrix $S \in \mathbb{R}^{m \times |s|}$, where each column represents the embedding of a word at the corresponding position in the sentence. Matrix $S$ is passed through a convolution layer. 
In this layer, a set of $f$ filters is applied to a sliding window of length $h$ over $S$ to generate a feature map matrix $O$. Each feature map $o_i$ for a given filter $F$ is generated by $o_i = \sum_{k,j}S[i:i+h]_{k,j} F_{k,j}$, where $S[i:i+h]$ denotes the concatenation of word vectors from position $i$ to $i+h$. The concatenation of all $o_i$ produces a feature vector $o \in \mathbb{R}^{|s|-h+1}$. The vectors $o$ are then aggregated over all $f$ filters into a feature map matrix $O \in \mathbb{R}^{f\times(|s|-h+1)}$.

We also add a bias vector $b \in R^f$ to the result of a convolution.
Each convolutional layer is followed by a non-linear activation function (we use ReLU) which is applied element-wise. Afterward, the output is passed to the max pooling layer which operates on columns of the feature map matrix $O$ returning the largest value: $pool(o_i) : \mathbb{R}^{1\times(|s|-h+1)} \rightarrow \mathbb{R}$ (see Figure~\ref{fig:sentiment}). 
This architecture is similar to the state-of-the-art model for Twitter sentiment classification from Semeval 2015 and 2016~\citep{Severyn:2015:SemEval,Deriu2016:SemEval}.

We initialize the embedding matrix with word2vec embeddings
%~\citep{Mikolov:2013} 
pretrained on a collection of 50M tweets.

\mypar{The Supervision Layer} receives the vector representation of the inputs processed by the representation learning layer and outputs a prediction $\tilde{y}$.
We opt for a simple fully connected feed-forward network with $l$ hidden layers followed by a softmax. 
Each hidden layer $z_k$ in this network computes $z_k = \alpha(W_k z_{k-1} + b_k)$, where $W_k$ and $b_k$ denote the weight matrix and the bias term corresponding to the $k^{th}$ hidden layer and $\alpha(.)$ is the non-linearity. These layers follow a softmax layer which returns $\tilde{y}_i$, the probability distribution over all three classes. 
We employ the weighted cross entropy loss:
\begin{equation}
% \nonumber
\mathcal{L}_t = \sum_{i\in B_U} \tilde{c}_i \sum_{k \in K} - \tilde{y}_i^k \log (\hat{y}_i^k),
\end{equation}
where $B_U$ is a batch of instances from $U$, and $\tilde{c}_i$ is the confidence score of the weakly annotated instance $i$, and $K$ is a set of classes.

\mypar{The Weak Annotator}\label{sentiment-WA} for the sentiment classification task is a simple unsupervised lexicon-based method~\citep{Hamdan:2013,Kiritchenko:2014}.
We use SentiWordNet03~\citep{Gaccianella:2010} to assign probabilities (positive, negative and neutral) for each token in set $U$.
Then a sentence-level distribution is derived by simply averaging the distributions of the terms, yielding a noisy label $\tilde{y}_i \in \mathbb{R}^{|K|}$, where $|K|$ is the number of classes, i.e. $|K|=3$. 
We empirically found that using soft labels from the weak annotator works better than assigning a single hard label.
The target label $c_j$ for the \cnet is calculated by using the mean absolute difference of the true label and the weak label: $c_j= 1-\frac{1}{|K|}\sum_{k\in K}|y_j^k - \tilde{y}_j^k|$, where $y_j$ is the one-hot encoding of the sentence label over all classes.

\section{Experimental Setups}
\label{app:setup}
The proposed architectures are implemented in TensorFlow~\citep{tang2016:tflearn,tensorflow2015-whitepaper}. We use the Adam optimizer~\citep{Kingma:2014} and the back-propagation algorithm. Furthermore, to prevent feature co-adaptation, we use \emph{dropout}~\citep{Srivastava:2014} as a regularization technique in all models.

In our setup, the \cnet to predict $\tilde{c}_j$ is a fully connected feed forward network.
Given that the \cnet is learned only from a small set of true labels and to speed up training we initialize the representation learning layer with pre-trained parameters, i.e., pre-trained word embeddings.
We use ReLU as a non-linear activation function $\alpha$ in both \tnet and \cnet.

\mypar{Collections.}
We test our model on the twitter message-level sentiment classification of SemEval-15 Task 10B \citep{rosenthal:2015}. Datasets of SemEval-15 subsume the test sets from previous editions of SemEval, i.e. SemEval-13 and SemEval-14. Each tweet was preprocessed so that URLs and usernames are masked.

\mypar{Data with true labels.} 
We use train (9,728 tweets) and development (1,654 tweets) data from SemEval-13 for training and SemEval-13-test (3,813 tweets) for validation.
To make our results comparable to the official runs on SemEval we use SemEval-14 (1,853 tweets) and  SemEval-15 (2,390 tweets) as test sets~\citep{rosenthal:2015, Nakov:2016}.

\mypar{Data with weak labels.}
We use a large corpus containing 50M tweets collected during two months for both, training the word embeddings and creating the weakly annotated set $U$ using the lexicon based method explained in Section~\ref{sentiment-WA}. 

\mypar{Parameters and Settings.}
We tuned hyper-parameters for each model, including baselines, separately with respect to the true labels of the validation set using batched GP bandits with an expected improvement acquisition function~\citep{Desautels:2014}.  
The size and number of hidden layers for the classifier and the \cnet were separately selected from $\{32, 64, 128\}$ and $\{1, 2, 3\}$, respectively.
We tested the model with both, $1$ and $2$ convolutional layers. The number of convolutional feature maps and the filter width is selected from $\{200,300\}$ and $\{ 3, 4, 5\}$, respectively. The initial learning rate and the dropout parameter were selected from $\{1E-3, 1E-5\}$ and $\{0.0, 0.2, 0.5\}$, respectively. We considered embedding sizes of $\{100, 200\}$ and the batch size in these experiments was set to $64$.

\end{document}